\documentstyle[eaclap]{article}
\author{Rani Nelken \\
Tel-Aviv University \\ Tel-Aviv 69978, Israel \\
{\tt nelken@math.tau.ac.il} \And
Nissim Francez\\
Computer Science Department\\  The Technion\\ Haifa 32000, Israel \\
{\tt francez@cs.technion.ac.il}}
\title{Splitting the Reference Time:
Temporal Anaphora and Quantification in DRT}
\begin{document}

\maketitle
\vspace{-0.5in}
\begin{abstract}
This paper presents
an analysis of temporal anaphora in sentences which contain
quantification over events, within the framework of Discourse Representation
Theory. The analysis in \cite{Partee84} of quantified sentences, introduced
by a temporal connective, gives the wrong truth-conditions when the temporal
connective in the subordinate clause is  {\em before} or {\em after}.
This problem has been previously analyzed in \cite{deSwart} as an
instance of the {\em proportion problem}, and given a solution from a
Generalized Quantifier approach.
By using a careful distinction between the different notions of
{\em reference time}, based on \cite{FDTL}, we propose a solution to
this problem, within the framework of DRT. We show some applications of this
solution to additional temporal anaphora phenomena in quantified sentences.

\end{abstract}

\bibliographystyle{acl}

%% Macros
\newcommand{\hl}[1]{\rule{#1}{.8pt}}

%DRT macros - adapted from Uwe Reyle

\newcommand{\abut}{$\supset \hspace{-0.5ex} \subset$}

\def\drs#1#2{\begin{tabular}{|c|}\hline #1 \\ \\
		[-8pt] #2\\[-8pt] \\ \hline \end{tabular} }

\def\imp#1#2#3#4{\drs{#1}{#2} \ $\Rightarrow$ \ \drs{#3}{#4}}

\def\dis#1#2#3#4{\drs{#1}{#2} \ $\vee$ \ \drs{#3}{#4}}

\def\dup#1#2#3#4#5#6{\begin{tabular}{c}
			\drs{#1}{#2}\hspace{-.3 em}%

                        \raute{#4}{#3}% \hspace{-0.6 em}%
                        \drs{#5}{#6} \\
		     \end{tabular}}

\def\raute#1#2{\setlength{\unitlength}{2 em}
	\begin{picture}(2,2)(0,0)
	\put (1,1){\line(1,-1){1}}
	\put (1,1){\line(-1,-1){1}}
	\put (0,0){\line(1,-1){1}}
	\put (1,-1){\line(1,1){1}}
	\put (1,0.15){\makebox(0,0)[b]{#1}}
	\put (1,-0.5){\makebox(0,0)[b]{#2}}
	\end{picture}}

\def\topdrs#1#2{\begin{tabular}[t]{|c|}\hline #1 \\ \\
		[-8pt] #2\\[-8pt] \\ \hline \end{tabular} }

% Macros for examples - adapted from Micheal Covington
\newcommand{\exampleno}{\refstepcounter{equation}\theequation}
\newenvironment{example}{\begin{examples}\item}{\end{examples}}
\newcounter{equationsave}
\newenvironment{examples}
{%
\begin{list}{(\theequation)}
{%
\setcounter{equationsave}{\arabic{equation}}%
\usecounter{equation}%
\setcounter{equation}{\arabic{equationsave}}%
\setlength{\listparindent}{0pt}%
\def\makelabel##1{##1\hfil}%
}%
\raggedright}%
{\end{list}}

\section{Introduction}
The analysis of temporal expressions in natural language discourse provides a
challenge for contemporary semantic theories. \cite{Partee73} introduced the
notion of {\em temporal anaphora}, to account for ways in which temporal
expressions depend on surrounding elements in the discourse for their semantic
contribution to the discourse. In this paper, we discuss the interaction of
temporal anaphora and quantification over eventualities. Such interaction,
while interesting in its own right, is also a good test-bed for theories of the
semantic interpretation of temporal expressions. We discuss cases such as:

\begin{examples}
	\item {\sf {\em Before} John makes a phone call, he {\em always}
	    lights up a cigarette.} \cite{Partee84}
	    \label{sentence:always}

	\item {\sf {\em Often}, {\em when} Anne came home late, Paul had
	    already prepared dinner.} \cite{deSwart}
	    \label{sentence:perfect}

	\item {\sf {\em When} he came home, he {\em always} switched on the
	    tv. He took a beer and sat down in his armchair to forget
	    the day.} \cite{deSwart}
	    \label{discourse:progression}

	\item {\sf {\em When} John is at the beach, he {\em always} squints
	    {\em when} the sun is shining.} \cite{deSwart}
	    \label{sentence:iteration}
\end{examples}

The analysis of sentences such as~(\ref{sentence:always}) in \cite{Partee84},
within the framework of {\em Discourse Representation Theory} (DRT) \cite{DRT}
gives the wrong truth-conditions, when the temporal connective in the
sentence is {\em before} or {\em after}. In DRT, such sentences trigger
box-splitting with the eventuality of the subordinate clause and an updated
reference time in the antecedent box, and the eventuality of the main clause
in the consequent box, causing undesirable universal quantification over the
reference time.

This problem is analyzed in \cite{deSwart} as an instance of the
{\em proportion problem} and given a solution from a Generalized Quantifier
approach. We were led to seek a solution for this problem {\em within} DRT,
because of DRT's advantages as a general theory of discourse, and its
choice as the underlying formalism in another research project of ours, which
deals with sentences such as~\ref{sentence:always}--\ref{sentence:iteration},
in the context of natural language specifications of computerized systems.
In this paper, we propose such a solution, based on a careful distinction
between different roles of Reichenbach's {\em reference time}
\cite{Reichenbach}, adapted from \cite{FDTL}.
Figure~\ref{drs:minpair} shows a `minimal pair' of DRS's for
sentence~\ref{sentence:always}, one according to Partee's\shortcite{Partee84}
analysis and one according to ours.

\begin{figure*}[htb]
\centering
\framebox[\textwidth]{
\mbox {
\drs{$n \, x \, r_0$}{John($x$)\\
     \imp{$e_1  \, r_1$}{$e_1 \subseteq r_0 \; \; r_1 < e_1$\\
			$e_1$:\framebox[2.0cm]{$x$ phone}}%make  a  phone call
	 {$e_2$}{$e_2 \subseteq r_1$\\
		 $e_2$:\framebox[2.2cm]{$x$ light up}}
}}
\drs{$n \, x \, s$}{John($x$) $n \subseteq s$\\
     $s$:\drs{}{\imp{$e \, t$}{$t=loc(e)$\\
			     $e$: \framebox[1.5cm]{$x$ phone}}%make a phone call
		  {$e' \, t'$}{$t' < t \: e' \subseteq t'$\\
		   $e'$: \framebox[1.8cm]{$x$ light up}}}}%a cigarette
}
\caption{a:Partee's analysis \hspace{5cm} b:Our analysis}
\label{drs:minpair}
\end{figure*}

\section{Background}
An analysis of the mechanism of temporal anaphoric reference hinges upon an
understanding of the ontological and logical foundations of temporal
reference. Different concepts have been used in the literature as primitives.
These range from temporal instants in {\em Tense logic} \cite{Prior},
through intervals of time \cite{BennetPartee} as in the analysis of temporal
connectives in \cite{Heinamaki}, to event structures \cite{Kamp79} as in
Hinrichs' \shortcite{Hinrichs86} analysis of temporal anaphora.

An important factor in the interpretation of temporal expressions is the
classification of situations into different aspectual classes (or
Aktionsarten), which is based on distributional and semantic properties.
In this paper, we only consider {\em events} and {\em states}, together termed
{\em eventualities} in \cite{Bach}. In narrative sequences, event clauses seem
to advance the narrative time, while states block its progression. The
mechanism used to account for this phenomena in \cite{Hinrichs86} and
\cite{Partee84}, is based on the notion of {\em reference time}, originally
proposed by Reichenbach \shortcite{Reichenbach}.

Reichenbach's well - known account of the interpretation of the different
tense forms uses the temporal relations between three temporal indices: the
{\em utterance time}, {\em event time} and {\em reference time}. The reference
time according to \cite{Reichenbach} is determined either by context, or by
temporal adverbials.

\subsection{A unified analysis of temporal anaphora}
Hinrichs' and Partee's use of a notion of reference time, provides for
a unified treatment of temporal anaphoric relations in discourse, which
include narrative progression especially in sequences of simple past tense
sentences, temporal adverbs and temporal adverbial clauses, introduced by a
temporal connective. This concept of reference time is no longer an instant
of time, but rather, an interval. This approach can be summarized as follows:
in the processing of a discourse, the discourse-initial sentence is argued
to require some contextually determined reference time. Further event clauses
in the discourse introduce a new event, which is included within the
then-current reference time. Each such event also causes the reference time
to be updated to a time `just after' \cite{Partee84} this event. State
clauses introduce new states, which include the current reference time, and
do not update it.

As an example of such an analysis consider the following narrative discourse
\cite{Partee84}:

\begin{example}
{\sf John got up, went to the window, and raised the blind. It was light out.
He pulled the blind down and went back to bed. He wasn't ready to face the
day. He was too depressed.}
\end{example}

Figure~\ref{drs:Hinrichs narrative} shows a DRS for the first two sentences
of this discourse, according to Hinrichs' and Partee's analysis. The `$n$' in
the top DRS is a mnemonic for `now'- the utterance time. The first event in
the discourse, $e_1$ -- John's getting up -- is interpreted relative to a
contextually understood reference time, $r_0$. The event $e_1$ is included in
the current reference time, $r_0$. A new reference time marker, $r_1$ is then
introduced. $r_1$ lies immediately after $r_0$ (recorded as
$r_0 \preceq r_1$). $r_1$ serves as the current reference time for the
following event $e_2$. We continue in this fashion, updating the reference
time, until the second sentence in the discourse is processed. This sentence
denotes a state, $s_1$, which includes the then-current reference time.

\begin{figure}[htb]
\centering
\drs{$r_0 \, e_1 \, r_1 \, e_2 \, r_2 \, e_3 \, s_1 \, n$}{
$r_0 < n \: e_1 \subseteq r_0 \: r_0 \preceq r_1 $\\
$r_1 < n \: e_2 \subseteq r_1 \: r_1 \preceq r_2$\\
$r_2 < n \: e_3 \subseteq r_2 \: r_2 \subseteq s_1$\\
$e_1$: \framebox[3.0cm]{John get up}\\
$e_2$: \makebox[3.0cm][l]{$\ldots$}
}
\caption{}
\label{drs:Hinrichs narrative}
\end{figure}

Adverbial phrases, whether phrasal (e.g. {\sf `On Sunday'}) or clausal
(e.g. {\sf `When Bill left'}), are processed before the main clause. They
introduce a reference time, which overrides the current reference time,
and provides an anaphoric antecedent for the tense in the main clause.
This mechanism is used to explain how tense and temporal adverbials can
combine to temporally locate the occurrence, without running into problems
of relative scope \cite{Hinrichs88}. The tense morpheme of the main clause
locates the event time with respect to the reference time, whereas temporal
adverbials are used to locate the reference time.

{\em When}-clauses, for example, introduce a new reference time, which is
ordered after the events described in the preceding discourse. The
eventuality in the {\em when}-clause is related to this reference time as
discussed earlier with respect to narrative progression: a state includes its
reference time, while an event is included in it.
The eventuality in the main clause is interpreted with respect to this
reference time. If the main clause is an event-clause, this event
introduces a new reference time, just after the event time of the main
clause. As an example, consider the following discourse \cite{Partee84}:

\begin{example}
{\sf Mary turned the corner. When John saw her, she crossed the street. She
hurried into a store.}
\end{example}

Following Partee \shortcite{Partee84}, we will not construct a full DRS for
this discourse, but illustrate it with a diagram in
Figure~\ref{figure:circles}, with circles denoting inclusion.

\begin{figure}
\hl{7.3cm}
\centering
\setlength{\unitlength}{0.006500in}
\begin{picture}(400,92)(45,725)
\thinlines
\put( 75,755){\circle{60}}
\put(300,755){\circle{58}}
\put(386,757){\circle{58}}
\put( 75,805){\makebox(0,0)[lb]{$r_0$}}
\put(290,805){\makebox(0,0)[lb]{$r_2$}}
\put( 57,750){\makebox(0,0)[lb]{$e_{turn}$}}
\put(140,750){\makebox(0,0)[lb]{$r_1$}}
\put(195,750){\makebox(0,0)[lb]{$e_{see}$}}
\put(277,750){\makebox(0,0)[lb]{$e_{cross}$}}
\put(115,750){\makebox(0,0)[lb]{$\leq$}}
\put(165,750){\makebox(0,0)[lb]{$<$}}
\put(250,750){\makebox(0,0)[lb]{$\leq$}}
\put(335,750){\makebox(0,0)[lb]{$\leq$}}
\put(425,750){\makebox(0,0)[lb]{$\leq$}}
\put(445,750){\makebox(0,0)[lb]{$r_4$}}
\put(360,750){\makebox(0,0)[lb]{$e_{hurry}$}}
\put(380,805){\makebox(0,0)[lb]{$r_3$}}
\end{picture}
\hl{7.3cm}
\caption{}
\label{figure:circles}
\end{figure}

\subsection{Quantification over events}
\cite{Partee84} extends Hinrichs' treatment of temporal anaphora to the
analysis of sentences, which contain a temporal adverbial and quantification
over eventualities. According to her analysis, these trigger box-splitting as
do {\em if} or {\em every} clauses in DRT \cite{DRT}. Consider the following
example from \cite{Partee84}:

\begin{example}
  {\sf Whenever Mary telephoned, Sam was asleep.}
\label{sentence:whenever}
\end{example}

\begin{figure}[htb]
\centering
\drs{\topdrs{$n \, x \, y \, r_0$}{Mary($y$)\\
				   Sam($x$)}\\
			$\swarrow$}
    {\imp{$e_1 \, r_1$}{$e_1 \subseteq r_0 \: e_1 < n$\\
			$e_1 \preceq r_1 \: r_1 < n$\\
			$e_1$: \framebox[2.0cm]{$y$ telephone}
			}
	 {$s_1$}{$r_1 \subseteq s_1$\\
		 $s_1$: \framebox[2.0cm]{$x$ sleep}}}
\caption{}
\label{drs:whenever}
\end{figure}

The subordinate clause cannot be interpreted relative to a single reference
time, since Mary's telephoning is not specified
to occur at some specific time. Still, the sentence needs to be interpreted
relative to a reference time. This reference time can be a large interval, and
should contain each of the relevant occurrences of Mary's telephoning during
which Bill was asleep. This reference time is represented as $r_0$ in the top
sub-DRS.

The {\em `whenever'} triggers box-splitting. The event marker - $e_1$ is
introduced in the antecedent box, with the condition that it be temporally
included in the current reference time, $r_0$ and be prior to $n$.
The {\em `whenever'} also causes the introduction of $r_1$, a new reference
time marker. $r_1$ lies `just after' $e_1$. The stative clause causes the
introduction of $s_1$, which includes the reference time $r_1$.

The embedding conditions for the whole construction are just like those for a
regular {\em `if'} or {\em `every'} clause, i.e. the sentence is true, if
every proper embedding of the antecedent box can be extended to a proper
embedding of the combination of the antecedent and the consequent boxes.
This means, as desired, that for each choice of an event $e_1$ of Mary's
telephoning, and reference time $r_1$ `just after' it, there is a state of
Sam's being asleep, that surrounds $r_1$.

A sentence such as~(\ref{sentence:whenever}a) which is the same as
sentence~\ref{sentence:whenever}, except the {\em `whenever'} is replaced by
{\em `when'}, and {\em `always'} is added in the main clause,
would get the same DRS.

\begin{flushleft}
(\ref{sentence:whenever}a) {\sf When Mary telephoned, Sam was
always asleep.}
\end{flushleft}

\subsection{Extending the analysis}
As noted in \cite{Partee84}, this analysis does not extend in a
straightforward manner to cases in which the operator {\em when} is replaced
by (an unrestricted) {\em before} or {\em after}, in such quantified contexts.
Constructing a similar DRS for such sentences gives the wrong truth
conditions. For example, Figure~\ref{drs:minpair}a shows a DRS for
sentence~\ref{sentence:always}, according to the principles above.
$r_1$~-~the reference time, used for the interpretation of the main clause is
placed in the universe of the antecedent box. Because the temporal connective
is `before', $r_1$ is restricted to lie before $e_1$. The embedding conditions
determine, that this reference time be universally quantified over, causing an
erroneous reading in which for each event, $e_1$, of John's calling, for each
earlier time $r_1$, he lights up a cigarette. Paraphrasing this, we could say
that John lights up cigarettes at all times preceding each phone call, not
just once preceding each phone call. We did not encounter this problem in the
DRS in Figure~\ref{drs:whenever}, since although the reference time $r_1$,
is universally quantified over in that DRS as well, it is also restricted, to
immediately follow $e_1$. It is similarly restricted if `before' is replaced
with `just before' or `ten minutes before'. But, (unrestricted) `before' is
analyzed as `some time before', and thus the problem arises. We will
henceforth informally refer to this problem as Partee's quantification problem.

Partee \shortcite{Partee84} suggests that in these cases we somehow have to
insure that the reference time, $r_1$, appears in the universe of the
consequent DRS, causing it to be existentially quantified over, giving the
desired interpretation. De Swart \shortcite{deSwart} notes that simply moving
$r_1$ to the right-hand box does not agree with Hinrichs' assumption, that
temporal clauses are processed before the main clause, since they update the
reference time, with respect to which the main clause will be interpreted.
In our proposed solution, the `reference time' is indeed moved to the right
box, but it is a different notion of reference time, and (as will be shown)
exempt from this criticism.

\section{The proportion problem}
De Swart \shortcite{deSwart} sees Partee's quantification problem as a
temporal manifestation of the {\em proportion problem}, which arises in cases
such as \cite{Kadmon}:

\begin{example}
     {\sf Most women who own a cat are happy.}
\end{example}

The sentence is false in the case where out of ten women, one owns 50 cats
and is happy, while the other nine women own only one cat each, and are
miserable. This will not be predicted by the unselective binding of
quantifiers in DRT, which quantify over all the free variables in their
scope, in this case women-cat pairs. According to \cite{deSwart} Partee's
quantification problem is similar - the universal quantifier in sentences such
as~(\ref{sentence:always}) binds pairs of events and updated reference times,
where the desired quantificational scheme is universal quantification for the
event and existential for the reference time.

De Swart \shortcite{deSwart} offers a solution from a Generalized
Quantifier approach, based on the analysis of quantified NPs in transitive
sentences. In this analysis, the reference time is an implicit variable,
which is needed in the interpretation of the temporal relation, but is not
part of the quantificational structure.

Temporal connectives are viewed as relations, $TC$, between two sets of events:

\begin{example}
$\{<e_1,e_2> | <e_1,e_2> \in TC\}$
\label{TC}
\end{example}

The quantificational structure of such sentences can be analyzed either by an
iteration of monadic quantifiers, or as a single dyadic quantifier of
type $<1,1,2>$. In the first approach, adverbs of quantification (Q-adverbs)
are assigned the structure:

\begin{example}
$Q(S_s,\{e_1|\exists(S_m,TC_{e_1})\})$
\label{Q:iterated}
\end{example}

In~\ref{Q:iterated}, $S_s$ and $S_m$ denote, respectively, the sets of events
described by the subordinate and the main clause, $TC_{e_1}$ denotes the image
set of $e_1$ under the temporal connective $TC$, i.e. the set of events $e_2$
which are related to $e_1$ via the relation $TC$, (presented in~\ref{TC}).
In the second approach, the structure is:

\begin{example}
$[Q,\exists](S_s,S_m,TC)$
\label{Q:dyadic}
\end{example}

De Swart's solution does overcome Partee's quantification problem, although not
within DRT. As such, the existential quantification in~\ref{Q:dyadic} has to
be stipulated, whereas our analysis acquires this existential quantification
`for free'.

\section{Splitting the role of reference time}
Our analysis of Partee's quantification problem uses a different notion of
reference time than that used by the accounts in the exposition above.
Following \cite{FDTL}, we split the role of the reference time, used to
account for a large array of phenomena, into several independent mechanisms.
This separation allows for an analysis in DRT of temporal subordinate clauses
in quantified sentences, which avoids Partee's problem altogether. The
mechanisms we discuss are: the {\bf location time}, {\bf Rpt}
and {\bf perf}\footnote{An additional mechanism is the {\bf TPpt}, which for
simplicity's sake  will not be discussed in this paper.}. DRSs will contain
temporal markers corresponding to location times and Rpts.

The location time is an interval, used to temporally locate eventualities,
in accordance with their aspectual classification. Events are included in
their location time (recorded in the DRS as $e \subseteq t$ on the respective
markers), while states temporally overlap their location time (recorded as
$s \bigcirc t$). The verb tense determines the relation between the location
time and the utterance time e.g. if the tense is simple past, the location
time lies anteriorly to the utterance time. When it is simple present, the
location time coincides with the utterance
time\footnote{Since the utterance time, $n$ is a point in \cite{FDTL}, the
overlap relation between a state that holds in the present and $n$ reduces to
inclusion.}. Temporal adverbials restrict the location time: temporal adverbs
introduce a DRS-condition on the location time, while temporal subordinate
clauses introduce a relation between the event
time\footnote{The event time $t$ of an eventuality $e$ is the smallest
interval which includes $e$ (recorded as $t=loc(e)$).} of the subordinate
clause and the location time of the main clause. The exact temporal relation
denoted by a temporal connective depends on the aspectual classes of the
eventualities related by it\footnote{For the sake of the current presentation,
we assume the following relations for {\em When}: if both the
{\em when}-clause and the main clause denote states, then their respective
time indices overlap. If both are events then the times are temporally close,
with the exact relation undetermined. When one is a state and one an event,
then the time index of the state includes that of the event cf.
\cite{Hinrichs86}.}. For example, in the following
sentence~\ref{sentence:location time}, the event triggers the
introduction of an event marker $e$, and location time marker $t$, into the
DRS with the DRS-condition $e \subseteq t$. The past tense of the verb adds
the condition $t<n$.
In sentence~\ref{sentence:location time when}, the location time of the event
in the main clause is restricted to fall (just) after the event time of the
event of the subordinate clause.

\begin{examples}
\item {\sf Mary wrote the letter.}
	\label{sentence:location time}
\item {\sf Mary wrote the letter when Bill left.}
	\label{sentence:location time when}
\end{examples}

Narrative progression is dealt with by using the feature Rpt (or reference
point). The Rpt can be either an event or a time discourse marker, already
present in the DRS (recorded as assignment $Rpt:=e$). Eventualities are
interpreted with respect to the Rpt - events are taken to follow the current
Rpt, while states include it. The Rpt is reset during the processing of the
discourse. Note that in a `terminal' DRS (ready for an embedding test), all
the auxiliary Rpts `disappear' (do not participate in the embedding).

The perfect is analyzed by using the notion of a {\em nucleus}
\cite{MoensSteedman} to account for the inner structure of an eventuality.
A nucleus is defined as a structure containing a {\em preparatory process},
{\em culmination} and {\em consequent state}.
The categorization of verb phrases into different aspectual classes can be
phrased in terms of which part of the nucleus they refer to. The perfect is
seen in \cite{FDTL} as an aspectual operator. The eventualities described
by the perfect of a verb refer to the consequent state of its nucleus.
For example, the following sentence~\ref{sentence:present perfect} denotes the
state, $s$, holding at the present, that Mary has met the president. This
state is a result of the event $e$, in which Mary met the president.
Temporally, the state $s$ starts just when $e$ ends, or as it is put in
\cite{FDTL}:$e$ and $s$ {\em abut}, (represented as $e$ \abut $s$).

\begin{example}
	{\sf Mary {\em has met} the president.}
\label{sentence:present perfect}
\end{example}

\section{An alternative solution}
By extending the analysis of temporal subordinate clauses in \cite{FDTL},
to sentences which include quantification over eventualities, we can propose
an alternative DRT solution to Partee's quantification problem.
As in \cite{Partee84}, such sentences trigger box-splitting.
But now, the location time of the eventuality in the subordinate clause
serves as the antecedent for the location time of the eventuality in the
main clause. In this approach, each of the relevant temporal markers resides
in its appropriate box, yielding the correct quantificational structure. This
quantification structure does not need to be stipulated as part of the
Q-adverb's meaning, but arises directly from the temporal system.
We illustrate this analysis by constructing a DRS in Figure~\ref{drs:minpair}b
for sentence~\ref{sentence:always}.

In this DRS, $n$ denotes the utterance time. The subordinate clause triggers
the introduction of an event marker, $e$, with its event time marker $t$.
The main clause triggers the introduction of an event marker $e'$, and its
location time marker $t'$, with the DRS-condition $e' \subseteq t'$.
The assymetry in using the event time for $e$ and the location time for $e'$
arises from the interpretation rules of temporal connectives (for both
quantified and non-quantified sentences). Since the temporal connective in
this sentence is {\em before}, the relation between these two markers is one
of precedence.

We adopt a suggestion by Chierchia in \cite{Partee84}, that the whole
implication be rendered as a state. This state is no longer an atomic
eventuality. It is a complex state denoting John's habit. This state holds
during the present, and so its location time is $n$.

This solution is not prone to de Swart's \shortcite{deSwart} criticism against
the naive solution of moving the reference time to the right DRS. The temporal
clause may be processed before the main clause, since $t'$, the location time
of $e'$, which `replaces' $r_1$, the reference time of Partee's analysis, as
the temporal index of the eventuality in the the main clause, arises from
processing the main clause (not updating the reference time of the subordinate
clause).

\section{Additional phenomena}
In this section we present some applications of our analysis to related
constructions. First, we consider the past perfect, as in
sentence~\ref{sentence:perfect}. De Swart \shortcite{deSwart} gives this
example to illustrate the inability to interpret temporal connectives without
the use of the reference times. According to \cite{deSwart}, the subordinate
clause determines the reference time of the verb, which lies anteriorly to the
event time. Trying to use the event times would give the wrong analysis. This
would seem to be troublesome for our approach, which uses the location time of
the event in the main clause, and not its reference time. However, this is not
a problem, since our analysis of the perfect by the use of the operator
{\bf perf}, analyses the eventuality referred to by the main clause, as
the result state of a previous event. The temporal relation in the sentence
is inclusion between the event time of Anne's coming home, and the location
time of the result state of Paul's already having prepared dinner.

Next, we consider narrative progression in quantified contexts, as in
sentence~\ref{discourse:progression}. The basic construction is just the same
as in the paradigm structure, but now we have narrative progression in the
consequent box. This narrative progression is handled as ordinary narrative
progression in \cite{FDTL}, i.e. by resetting the Rpt. The DRS in
Figure~\ref{drs:narrative} describes the complex state $s_1$, that after each
event of John's coming home, there is a sequence of subsequent events
according to his activities.

\begin{figure}[htb]
\centering
\drs{$n \, x \, s_1 \, t_1$}{$s_1 \bigcirc t_1 \; \; t_1 < n$\\
     $s_1$:\drs{}{\imp{$e_1 \, t_2$}{$t_2 = loc(e_1)$\\
	      $e_1$:\framebox[1.2cm]{$x$ c.h.}}
	      {$e_2 \, e_3 \, e_4 \, t_3 \, t_4 \, t_5$}
	      {$e_2 \subseteq t_3 \: t_2 < t_3$\\
	       $e_2$:\framebox[1.8cm]{$x$ sw. on tv}\\
	       $Rpt:=e_2$\\
	       $e_3 \subseteq t_4 \: e_2 < e_3$\\
	       $e_3$: \framebox[1.8cm]{$x$ take beer}\\
	       $Rpt:=e_3$\\
	       $e_4 \subseteq t_4 \: e_3 < e_4$\\
	       $e_4$: \makebox[1.0cm][l]{$\ldots$}
	      	  }
		}
}
\caption{}
\label{drs:narrative}
\end{figure}

Finally, we deal with sentences such as (\ref{sentence:iteration}), which
contain an iteration of an implicit generic quantifier and {\em always}. The
situation described by John's always squinting when the sun is shining is
analyzed as a complex state $s_3$. This state holds whenever John is at the
beach, recorded by the condition that the location time $t_2$ of $s_3$
overlaps the event time, $t_1$ of John's being at the beach, $s_2$ in
Figure~\ref{drs:iteration}.

\begin{figure*}[htb]
\centering
\drs{$n \, x \, y \, s_1$}{John($x$) $\:$ the sun($y$) $\: n \subseteq s_1$\\
	   $s_1$:\drs{}{\imp{$s_2 \, t_1$}{$loc(s_2)=t_1$\\
			   $s_2$:\framebox[3.0cm]{$x$ is at the beach}\\
				          }
		   {$s_3 \, t_2$}{$s_3 \bigcirc t_2 \: t_2 \bigcirc t_1$\\
					  $s_3$:\drs{}{\imp{$t_3 \, s_4$}
			   {$loc(s_4)=t_3$\\
			    $s_4$:\framebox[2.5cm]{$y$ is shining}
			   }
			   {$e_1 \, t_4$}{$e_1 \subseteq t_4$\\
					  $t_4 \subseteq t_3$\\
					$e_1$:\framebox[2.0cm]{$x$ squint}}
						      }
					  }
		}
}
\caption{}
\label{drs:iteration}
\end{figure*}

\section{Acknowledgments}
The work of the second author was partially supported by a grant from the
Israeli ministry of science ``Programming languages induced computational
linguistics'', and by the fund for the promotion of research in the Technion.
The authors would like to thank Nirit Kadmon and Uwe Reyle for reading
a preliminary version of this paper.


\begin{thebibliography}{}

\bibitem[\protect\citename{Bach}1981]{Bach}
Emmon Bach.
\newblock 1981.
\newblock On time, tense and aspect. An essay in English metaphysics.
\newblock in Peter Cole (ed.), {\em Radical Pragmatics}
\newblock Academic Press, New York 63--81.

\bibitem[\protect\citename{Bennet and Partee}1978(1972)]{BennetPartee}
Michael Bennet and Barbara Partee.
\newblock 1978(1972).
\newblock Toward the Logic of Tense and Aspect in English
\newblock ms. Reproduced by the Indiana University Linguistics Club

\bibitem[\protect\citename{Hein\"am\"aki}1978]{Heinamaki}
Orvokki Hein\"am\"aki.
\newblock 1978.
\newblock {\em Semantics of English temporal connectives.}
\newblock Bloomington: Indiana University Linguistics Club.

\bibitem[\protect\citename{Hinrichs}1986]{Hinrichs86}
Erhard~W. Hinrichs.
\newblock 1986.
\newblock Temporal anaphora in discourses of English.
\newblock {\em Linguistics and Philosophy} 9:63--82.

\bibitem[\protect\citename{Hinrichs}1988]{Hinrichs88}
Erhard~W. Hinrichs.
\newblock 1988.
\newblock Tense, quantifiers and contexts.
\newblock {\em Computational Linguistics} volume 14, number 2, pages 3-14.

\bibitem[\protect\citename{Kadmon}1990]{Kadmon}
Nirit Kadmon.
\newblock 1990.
\newblock Uniqueness
\newblock {\em Linguistics and Philosophy} 13:273--324

\bibitem[\protect\citename{Kamp}1979]{Kamp79}
Hans Kamp.
\newblock 1979.
\newblock Events, instants and temporal reference.
\newblock In R. Ba\"uerle, U. Egli and A. von Stechow (eds.)
\newblock Semantics from different points of view
\newblock Springer Verlag, Berlin.

\bibitem[\protect\citename{Kamp}1981]{DRT}
Hans Kamp.
\newblock 1981.
\newblock A theory of Truth and Semantic  Representation
\newblock in J. Groenendijk, TH. Janssen and M. Stokhof(eds.)
\newblock Formal Methods in the Study of Language, Part I.
\newblock Mathematisch Centrum, Amsterdam. pages 277-322

\bibitem[\protect\citename{Kamp and Reyle}1993]{FDTL}
Hans Kamp and Uwe Reyle
\newblock 1993.
\newblock From Discourse to Logic {\em Introduction to Modeltheoretic
 Semantics of Natural Language, Formal Logic and Discourse Representation
 Theory.}
\newblock  Kluwer Academic Publishers, Dordrecht.

\bibitem[\protect\citename{Moens and Steedman}1988]{MoensSteedman}
Marc Moens and Mark Steedman
\newblock 1988.
\newblock Temporal Ontology and Temporal Reference
\newblock {\em Computational Linguistics} 14 15--28.

\bibitem[\protect\citename{Partee}1973]{Partee73}
Barbara~H. Partee.
\newblock 1973.
\newblock Some structural analogies between tenses and pronouns in English.
\newblock {\em Journal of Philosophy} LXX 601-609

\bibitem[\protect\citename{Partee}1984]{Partee84}
Barbara~H. Partee.
\newblock 1984.
\newblock Nominal and Temporal Anaphora
\newblock Linguistics and Philosophy 7.3 (243-286)

\bibitem[\protect\citename{Prior}67]{Prior}
Arthur Prior.
\newblock 1967.
\newblock Past, Present and Future.
\newblock Oxford University Press, Oxford

\bibitem[\protect\citename{Reichenbach}1947]{Reichenbach}
Hans Reichenbach.
\newblock 1947.
\newblock Elements of Symbolic Logic.
\newblock Reprinted in 1966 by The Free Press, New York.


\bibitem[\protect\citename{de Swart}1991]{deSwart}
Henri\"ette de Swart.
\newblock 1991.
\newblock Adverbs of Quantification: {\em a Generalized Quantifier Approach},
\newblock diss. University of Groningen. Published (1993) by Garland, New York.


\end{thebibliography}
\end{document}